\documentclass[letterpaper, 10 pt, conference]{ieeeconf}  %

\IEEEoverridecommandlockouts                              %

\usepackage{etex}
\usepackage{graphics} %
\usepackage{epsfig} %
\usepackage{newtxtext}
\usepackage{times} %
\usepackage{amsfonts} 
\usepackage{amsmath} %
\usepackage{amssymb}  %
\usepackage{bm}
\usepackage{float}
\usepackage{pgf}
\usepackage{hhline}
\usepackage{hyperref}
\usepackage{cite}
\usepackage{multirow}
\usepackage{colortbl}
\usepackage{array}
\usepackage{comment}
\usepackage{siunitx}
\usepackage{relsize}
\usepackage{ifthen}

\usepackage{subcaption}

\usepackage[vlined,ruled,linesnumbered]{algorithm2e}
\usepackage{graphics} %
\usepackage{rotating}
\usepackage{color}
\usepackage{enumerate}
\usepackage[T1]{fontenc}
\usepackage{psfrag}
\usepackage{epsfig} %
\usepackage{booktabs}
\usepackage{graphicx,url}
\usepackage{multirow}
\usepackage{array}
\usepackage{latexsym}
\usepackage{amsfonts}
\usepackage{amsmath}
\usepackage{amssymb}
\usepackage{xstring}
\usepackage[noend]{algorithmic}
\usepackage{multirow}
\usepackage{prettyref}
\usepackage{flexisym}
\usepackage{bigdelim}
\usepackage{breqn} %
\usepackage{listings}
\let\labelindent\relax
\usepackage{enumitem}
\usepackage{xspace}
\usepackage{bm}
\graphicspath{{./figures/}}
\usepackage{tikz}
\usetikzlibrary{matrix,calc}

\usepackage{mdwlist}

\let\itemize\undefined
\makecompactlist{itemize}{stditemize}

\newrefformat{prob}{Problem\,\ref{#1}}
\newrefformat{def}{Definition\,\ref{#1}}
\newrefformat{sec}{Section\,\ref{#1}}
\newrefformat{sub}{Section\,\ref{#1}}
\newrefformat{prop}{Proposition\,\ref{#1}}
\newrefformat{app}{Appendix\,\ref{#1}}
\newrefformat{alg}{Algorithm\,\ref{#1}}
\newrefformat{cor}{Corollary\,\ref{#1}}
\newrefformat{thm}{Theorem\,\ref{#1}}
\newrefformat{lem}{Lemma\,\ref{#1}}
\newrefformat{fig}{Fig.\,\ref{#1}}
\newrefformat{tab}{Table\,\ref{#1}}

\newcommand{\bdmath}{\begin{dmath}}
\newcommand{\edmath}{\end{dmath}}
\newcommand{\beq}{\begin{equation}}
\newcommand{\eeq}{\end{equation}}
\newcommand{\bdm}{\begin{displaymath}}
\newcommand{\edm}{\end{displaymath}}
\newcommand{\bea}{\begin{eqnarray}}
\newcommand{\eea}{\end{eqnarray}}
\newcommand{\beal}{\beq \begin{array}{ll}}
\newcommand{\eeal}{\end{array} \eeq}
\newcommand{\beas}{\begin{eqnarray*}}
\newcommand{\eeas}{\end{eqnarray*}}
\newcommand{\ba}{\begin{array}}
\newcommand{\ea}{\end{array}}
\newcommand{\bit}{\begin{itemize}}
\newcommand{\eit}{\end{itemize}}
\newcommand{\ben}{\begin{enumerate}}
\newcommand{\een}{\end{enumerate}}

\newcommand{\etal}{\emph{et~al.}\xspace}
\newcommand{\setal}{~\emph{et~al.}\xspace}
\newcommand{\M}[1]{{\bm #1}} %
\renewcommand{\boldsymbol}[1]{{\bm #1}}

\newcommand{\hide}[1]{}

\newcommand{\hiddenText}{{\color{gray} hidden text.}}
\newcommand{\hideWithText}[1]{\hiddenText}

\newcommand{\Real}[1]{ { {\mathbb R}^{#1} } }

\newcommand{\SE}[1]{\ensuremath{\mathrm{SE}(#1)}\xspace}

\newcommand{\SO}[1]{\ensuremath{\mathrm{SO}(#1)}\xspace}

\newcommand{\MR}{\M{R}}

\newcommand{\vt}{\boldsymbol{t}}
\newcommand{\vxx}{\boldsymbol{x}}

\newcommand{\scenario}[1]{{\smaller \sf#1}\xspace}

\newcommand{\blue}[1]{{\color{blue}#1}}

\newcommand{\linkToPdf}[1]{\href{#1}{\blue{(pdf)}}}
\newcommand{\linkToPpt}[1]{\href{#1}{\blue{(ppt)}}}
\newcommand{\linkToCode}[1]{\href{#1}{\blue{(code)}}}
\newcommand{\linkToWeb}[1]{\href{#1}{\blue{(web)}}}
\newcommand{\linkToVideo}[1]{\href{#1}{\blue{(video)}}}
\newcommand{\award}[1]{\xspace} %

\newcommand{\vz}{\boldsymbol{z}}

\newcommand{\DOORSLAM}{\scenario{DOOR-SLAM}}
\newcommand{\PCM}{\scenario{PCM}}
\newcommand{\RTABMAP}{\scenario{RTAB-Map}}
\newcommand{\BUZZ}{\scenario{Buzz}}
\newcommand{\ARGOS}{\scenario{ARGoS}}
\newcommand{\NETVLAD}{\scenario{NetVLAD}}
\newcommand{\KITTI}{\scenario{KITTI00}}
\newcommand{\KITTIs}{\scenario{KITTI}}

\newcommand{\OPENCV}{\scenario{OpenCV}}

\newcommand{\barR}{\bar{\MR}}

\newcommand{\bart}{\bar{\vt}}
\newcommand{\barz}{\bar{\vz}}

\newcommand{\modification}[1]{\textcolor{black}{#1}} 
\title{\LARGE \bf
DOOR-SLAM: Distributed, Online, and Outlier Resilient SLAM\\ for Robotic Teams
}

\author{Pierre-Yves Lajoie$^{1}$, Benjamin Ramtoula$^{1,3}$, Yun Chang$^{2}$, Luca Carlone$^{2}$, Giovanni Beltrame$^{1}$%
\thanks{ *This work was partially funded by the Natural Sciences and Engineering Research Council of Canada (NSERC), 
 the J.A. DeS\`eve Foundation, %
 ARL DCIST CRA W911NF-17-2-0181, and the DARPA ``Specification-guided and Capability-aware Autonomy for Long-endurance 
Situational Awareness in Subterranean Environments'' project.
 }%
\thanks{$^{1}$Department\,of\,Computer\,and\,Software\,Engineering, Polytechnique Montr\'eal, Montreal, Canada, %
{\tt\scriptsize\,\{pierre-yves.lajoie,benjamin.ramtoula,}\newline
{\tt\scriptsize giovanni.beltrame\}@polymtl.ca}}
\thanks{$^{2}$Laboratory for Information \& Decision Systems (LIDS), Massachusetts \newline 
Institute of Technology,\,Cambridge,\,USA,\,{\tt\scriptsize\,\{yunchang,lcarlone\}@mit.edu}}
\thanks{$^{3}$School of Engineering, \'Ecole Polytechnique F\'ed\'erale de Lausanne, Switzerland.
}
}

\begin{document}

\maketitle
\thispagestyle{empty}
\pagestyle{empty}

\begin{abstract}
  To achieve collaborative tasks, robots in a team need to have a
  shared understanding of the environment and their location within it.
  Distributed Simultaneous Localization and Mapping (SLAM)
  offers a practical solution to localize the robots without relying on an
  external positioning system (e.g. GPS) and with minimal information
  exchange.  Unfortunately, current distributed SLAM systems are vulnerable to
  perception outliers and therefore tend to use very conservative parameters
  for inter-robot place recognition. However, being too conservative comes at
  the cost of rejecting many valid loop closure candidates, which results in
  less accurate trajectory estimates.  This paper introduces \DOORSLAM, a
  fully distributed SLAM system with an outlier rejection mechanism that can
  work with less conservative parameters. \DOORSLAM is based on peer-to-peer
  communication and does not require full connectivity among the robots.
  \DOORSLAM includes two key modules: a pose graph optimizer
  combined with a distributed \emph{pairwise consistent measurement set
    maximization} algorithm to reject spurious inter-robot loop closures; and
  a distributed SLAM front-end that detects inter-robot loop closures without
  exchanging raw sensor data.  The system has been evaluated in simulations,
  benchmarking datasets, and field experiments, including tests in GPS-denied
  subterranean environments. \DOORSLAM produces more inter-robot loop
  closures, successfully rejects outliers, and results in accurate trajectory
  estimates, while requiring low communication bandwidth.  Full source code is
  available at \url{https://github.com/MISTLab/DOOR-SLAM.git}.
\end{abstract}


\section{Introduction}
\label{sec:introduction}

Multi-robot systems already constitute the backbone of many modern robotics
applications, from warehouse maintenance to self-driving cars, and have the
potential to impact other endeavors, including search \& rescue and planetary
exploration.
These applications involve a team of robots completing a coordinated task in
an unknown or partially known environment, and require the robots to have a
shared understanding of the environment and their location within it. While a
common practice is to circumvent this need by adding external localization
infrastructure (e.g., GPS, motion capture, geo-referenced markers), such a
solution is not always viable; for instance, when robots are deployed for cave
exploration or building inspection, the deployment of an external
infrastructure may be dangerous, expensive, or impractical. Therefore,
multi-robot SLAM solutions that can work without external
localization infrastructure and provide reliable situational awareness are
highly desirable.

Obtaining such a shared situational awareness is challenging since the sensor
data required for SLAM is distributed across the robots, and communicating raw
data may be slow (due to bandwidth constraints) or infeasible (due to limited
communication range).  For these reasons, current systems either rely on a
centralized and offline post-processing step~\cite{Mangelson18icra}, assume
all robots are always within communication range~\cite{cieslewski_efficient_2017-bow}, or
assume centralized pre-processing of the sensor data (e.g., to remove
outliers~\cite{Choudhary17ijrr-distributedPGO3D}).  We believe more flexible
solutions are necessary for a broader adoption of multi-robot technologies.
\modification{For instance, bandwidth issues can be mitigated by relying on local exchange of
processed data among the robots to collaboratively compute a SLAM solution.}

\begin{figure}[t]
\centering
\includegraphics[width=0.99\columnwidth,trim=0mm 0mm 0mm 0mm,clip]{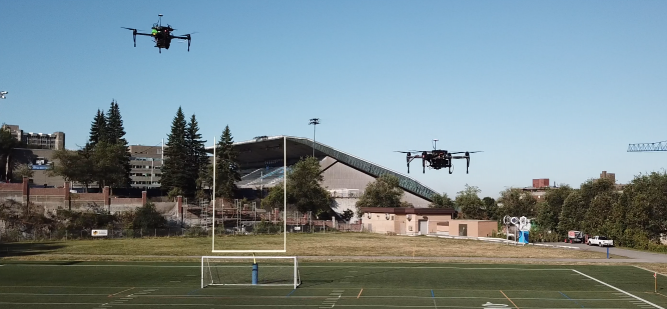}
\begin{subfigure}[b]{0.23\textwidth}
\centering
\includegraphics[width=\textwidth,trim=0mm 0mm 0mm 0mm,clip]{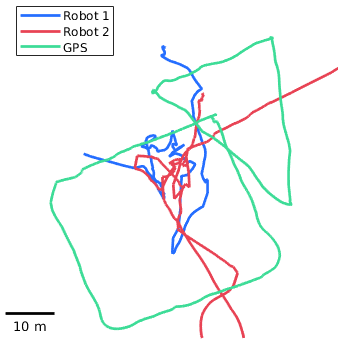}
\caption{\textbf{without} outlier rejection.}
\end{subfigure}
\begin{subfigure}[b]{0.23\textwidth}
\centering
\includegraphics[width=\textwidth,trim=0mm 0mm 0mm 0mm,clip]{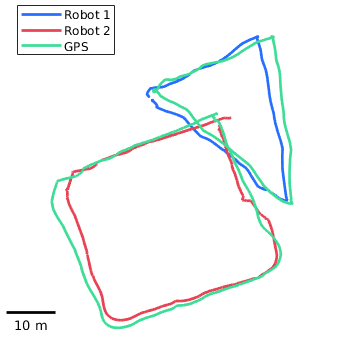}\caption{\textbf{with} outlier rejection.}
\end{subfigure}
\vspace{-2mm}
\caption{Trajectory estimates from \DOORSLAM
(red and blue) and GPS ground truth (green, only used for benchmarking).}
\label{fig:fieldExperimentsTrajectories}
\vspace{-6mm}
\end{figure}

In addition to the communication constraints, multi-robot SLAM is challenging
and prone to failures due to
incorrect data association and perceptual aliasing. The latter is particularly
problematic since it generates incorrect loop closures between scenes that
look similar but correspond to different places. While this topic has received
considerable attention in the centralized
case~\cite{Sunderhauf12iros,Agarwal13icra,Olson12rss,Latif14rss,Mangelson18icra,Lajoie19ral-DCGM},
the literature currently lacks \emph{distributed} outlier rejection methods.
We believe implementing distributed outlier rejection would improve the
robustness of multi-robot systems, allow users to be less conservative during
parameters tuning, and enable the detection of more loop closures, improving
the accuracy of the SLAM solution.

\textbf{Contribution.}  
In this system paper, we present \DOORSLAM, a fully distributed SLAM system for robotic teams.
\DOORSLAM has the following desirable features: (i) it does not require full connectivity maintenance between the robots, 
(ii) it is able to detect inter-robot loop closures without exchanging raw data, 
(iii) it performs distributed outlier rejection to remove incorrect inter-robot loop closures, and
(iv) it executes a distributed pose graph optimization to retrieve the robots' trajectory estimates.

The proposed system includes two key modules. The first module is a
pose graph optimizer that is robust to spurious measurements.
We propose an implementation of distributed pose graph optimization along the
lines of~\cite{Choudhary17ijrr-distributedPGO3D} combined with an outlier
rejection mechanism based on~\cite{Mangelson18icra}, that we adapted for
online and distributed operation. An example of the robustness afforded by
the proposed module is showcased in
Fig.~\ref{fig:fieldExperimentsTrajectories}, which reports the trajectory
estimates with and without outlier rejection.
Our implementation is robust to perceptual aliasing and allows practitioners
to use a less conservative tuning of the SLAM front-end.
The second module is a data-efficient distributed SLAM front-end. Similar to
the recent approach~\cite{Cieslewski18icra}, our system uses \NETVLAD
descriptors~\cite{Arandjelovic16cvpr-netvlad} for place recognition. \modification{However,
our approach trades off some data-efficiency to obviate full connectivity maintenance and
environment-specific pre-training requirements.}

\DOORSLAM has been evaluated in simulations, benchmarking datasets
(\scenario{KITTI}~\cite{Geiger12cvpr}), and field experiments, including tests
in GPS-denied subterranean environments.  \DOORSLAM runs online on an
{\smaller\sf NVIDIA Jetson TX2} computer, successfully rejects outliers, and
results in accurate trajectory estimates, while requiring a low bandwidth.
We release the source code and {\smaller\sf Docker} images for easy reuse of the system components by the community: \url{https://github.com/MISTLab/DOOR-SLAM.git}.   

\section{Related Work}
\label{sec:relatedWork}

\subsection{Distributed Pose Graph Optimization (PGO)}
\modification{Pose Graph Optimization (PGO) is a popular estimation engine
for SLAM.}  Centralized approaches for multi-robot PGO collect all measurements
at a central station, which computes the trajectory estimates for all the
robots~\cite{Andersson08icra,Kim10icra,Bailey11icra,Lazaro11icra,Dong15icra}.
\modification{Since the computation workload and the communication bandwidth of a
centralized approach grow with the number of robots, related work has also
explored \emph{distributed techniques}, in which robots only exploit local
computation and communication.
Aragues\setal~\cite{Aragues11icra-distributedLocalization} use a distributed
Jacobi approach to estimate 2D
poses. %
Cunningham\setal~\cite{Cunningham10iros,Cunningham13icra} use Gaussian elimination.}
\modification{Recent work from Choudhary~\etal~\cite{Choudhary17ijrr-distributedPGO3D} introduces the Distributed Gauss-Seidel approach, which supports 3D cases and avoids the complex bookkeeping and information double counting required by the previous techniques.}
\modification{It requires only to share the latest pose estimates involved in inter-robot measurements.}
Recent distributed SLAM
solutions~\cite{Cieslewski18icra} and~\cite{Wang19arxiv} have used the
implementation of Choudhary~\etal~\cite{Choudhary17ijrr-distributedPGO3D} as
back-end for their experiments. 
While here we focus on PGO, we refer the
reader to~\cite{Choudhary17ijrr-distributedPGO3D} for an extensive review on
other distributed estimation techniques.

\subsection{Robust PGO}
The problem of mitigating the effects of outliers in pose graph optimization
has received substantial attention in the literature, due to the dramatic
distortion that even one incorrect measurement can cause. Early work in the
field includes techniques such as RANSAC~\cite{Fischler81}, branch \&
bound~\cite{Neira01tra}, and M-estimation (see~\cite{Bosse17fnt,Hartley13ijcv}
for a review).  \modification{S\"underhauf~\etal~\cite{Sunderhauf12iros} introduce the idea of outliers deactivation
using binary variables that are then relaxed to continuous
variables.} \modification{Agarwal~\etal~\cite{Agarwal13icra} build on top of this idea to dynamically scale the
measurement covariances.} %
\modification{Other works on the single robot case include Olson and Agarwal~\cite{Olson12rss} and Pfingsthorn and
Birk~\cite{Pfingsthorn13ijrr,Pfingsthorn16ijrr} which consider multi-modal
distributions for the noise.
Recent work from
Lajoie~\etal~\cite{Lajoie19ral-DCGM} and Carlone and
Calafiore~\cite{Carlone18ral-robustPGO2D} focus on robust global solvers based
on convex relaxations.}
\modification{Instead of classifying the measurements individually, Latif~\etal~\cite{Latif14rss}, Carlone~\etal~\cite{Carlone14iros-robustPGO2D},
Graham~\etal~\cite{Graham15iros} look for sets of mutually consistent
measurements.}
\modification{Mangelson~\etal~\cite{Mangelson18icra} extend the latter idea to the
multi-robot case and propose an effective graph-theoretic technique to find
pairwise-consistent measurements among the inter-robot loop
closures.}
\modification{Alternatives for multi-robot cases include Dong~\etal~\cite{Dong15icra} which search for consistent inter-robot measurements using expectation maximization.}
\modification{Wang~\etal~\cite{Wang19arxiv} leverage extra information from wireless channels to detect outliers during a multi-robot rendezvous.}

\subsection{Distributed Loop Closure Detection}

Inter-robot loop closures are critical to align the trajectories of the robots
in a common reference frame and to improve the trajectory estimates.
In a centralized setup, a common way to obtain loop closures is to use visual
place recognition methods, which compare compact image descriptors to find
potential loop closures. This is traditionally done with global visual
features~\cite{Oliva01ijcv,Ulrich00icra}, or local visual
features~\cite{Lowe99iccv,Bay06eccv} which can be quantized in a bag-of-word
model~\cite{Sivic03iccv}. More recently, convolutional neural networks (CNN),
either using features trained on auxiliary
tasks~\cite{suenderhauf_performance_2015} or directly trained end-to-end for
place recognition, such as \NETVLAD~\cite{Arandjelovic16cvpr-netvlad}, have
generated more robust descriptors.  Geometric verification using local
features is then used to validate putative loop closures and estimate
transformations between the corresponding observation
poses~\cite{philbin_object_2007,Scaramuzza11ram}.

Distributed loop closure detection has the additional challenge that the
images are not collected at a single location and their exchange is
problematic due to range and bandwidth constraints.
Tardioli~\etal~\cite{Tardioli2015} use visual vocabulary indexes instead of descriptors to reduce the required bandwidth.
Cieslewski and Scaramuzza~\cite{Cieslewski18icra} propose distributed
and scalable solutions for place recognition in a fully connected team of
robots. A first approach~\cite{cieslewski_efficient_2017-bow} relies on
bag-of-words of visual features~\cite{Sivic03iccv} which are split and
distributed among the team. Another
one~\cite{cieslewski_efficient_2017-netvlad} pre-assigns a range of
descriptors from \NETVLAD to each robot, allowing place recognition search
over the full team by communicating with a single other robot. These methods
minimize the required bandwidth and scale well with the number of robots,
but are designed for situations with full connectivity in the team.
Tian~\etal~\cite{tian_near-optimal_2018,Tian2019} and
Giamou~\etal~\cite{giamou_talk_2017} propose complementary approaches to these
methods. They consider robots having rendezvous and efficiently coordinate the
data exchange during the geometric verification step, accounting for the available
communication and computation resources.

\section{The \DOORSLAM System}
\label{sec:methodology}

Our distributed SLAM system relies on peer-to-peer communication: each robot
performs single-robot SLAM when there is no teammate within communication
range, and executes a distributed SLAM protocol during a rendezvous.

Our implementation leverages \BUZZ~\cite{Pinciroli2016}, a
programming language specifically designed for multi-robot systems.  \BUZZ
offers useful primitives to build a fully decentralized software architecture,
and seamlessly handles the transition between single-robot and multi-robot
execution. \BUZZ is a scripting language that lets us abstract
away the details concerning communication, neighbor detection and management,
and provides a uniform framework to implement and compare multi-robot
algorithms (such as SLAM, task allocation, exploration, etc.). 
It provides a uniform gossip-based interface, implemented on WiFi,
Xbee, Bluetooth, or custom networking devices. \BUZZ is thought of as an
\emph{extension} language, i.e. it is designed to be laid on top of other
frameworks, such as the Robot Operating System (ROS). This allows us to run
\DOORSLAM on virtually any type and \emph{any number} of robots
that support ROS without modification. Experiments~\cite{Pinciroli2016} show that \BUZZ can scale up to thousands of robots.

\begin{figure}
\centering
  \includegraphics[width=\linewidth]{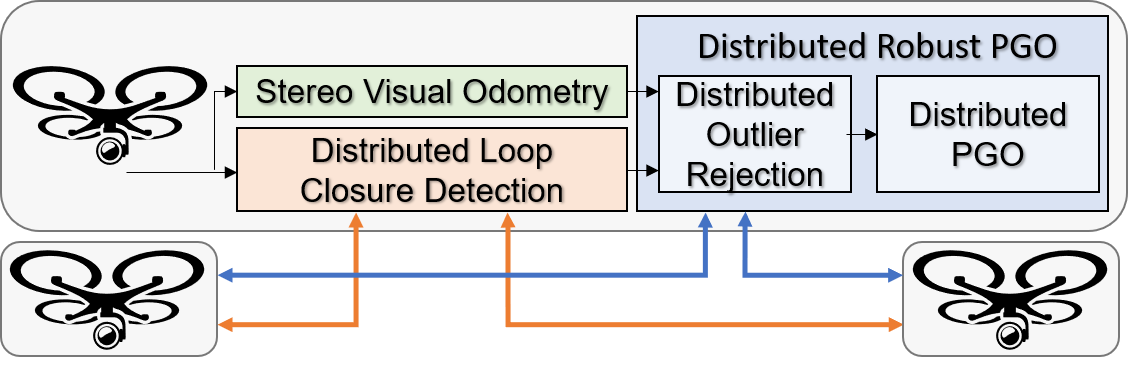} 
  \vspace{-5mm}
\caption{\DOORSLAM system overview}
\label{fig:systemOverview}
\vspace{-8mm}
\end{figure}

A system overview of \DOORSLAM is given in Fig.~\ref{fig:systemOverview}. Each
robot collects images from an onboard stereo camera and uses a (single-robot)
Stereo Visual Odometry module to produce an estimate of its trajectory. In our
implementation, we use the stereo odometry from \RTABMAP~\cite{Labbe2019}. The
images are also fed to the Distributed Loop Closure Detection module
(Section~\ref{sec:dist_loop_closure_detection}) which communicates information
with other robots (when they are within communication range) and outputs
inter-robot loop closure measurements. Then, the Distributed Outlier
Rejection module (Section~\ref{sec:dist_robust_pgo}) collects the odometry and
inter-robot measurements to compute the maximal set of pairwise consistent
measurements and filters out the outliers.  Finally, the Distributed Pose
Graph Optimization module (Section~\ref{sec:dist_robust_pgo}) performs
distributed SLAM.
For simplicity, in the current implementation, we only consider inter-robot
loop closures~\cite{Choudhary17ijrr-distributedPGO3D} (i.e., loop closures
involving poses of different robots).  The system can be easily extended to
use intra-robot loop closures (i.e., the loop closures commonly encountered in
single-robot SLAM) by replacing stereo odometry~\cite{Labbe2019} with a visual
SLAM solution.

In the following sections, we focus on the distributed place recognition
module and on the distributed robust PGO module, while we refer the reader
to~\cite{Labbe2019} for a description of the stereo visual odometry module.

\subsection{Distributed Loop Closure Detection}
\label{sec:dist_loop_closure_detection}

\begin{figure}[t!]
\centering
  \includegraphics[width=\linewidth]{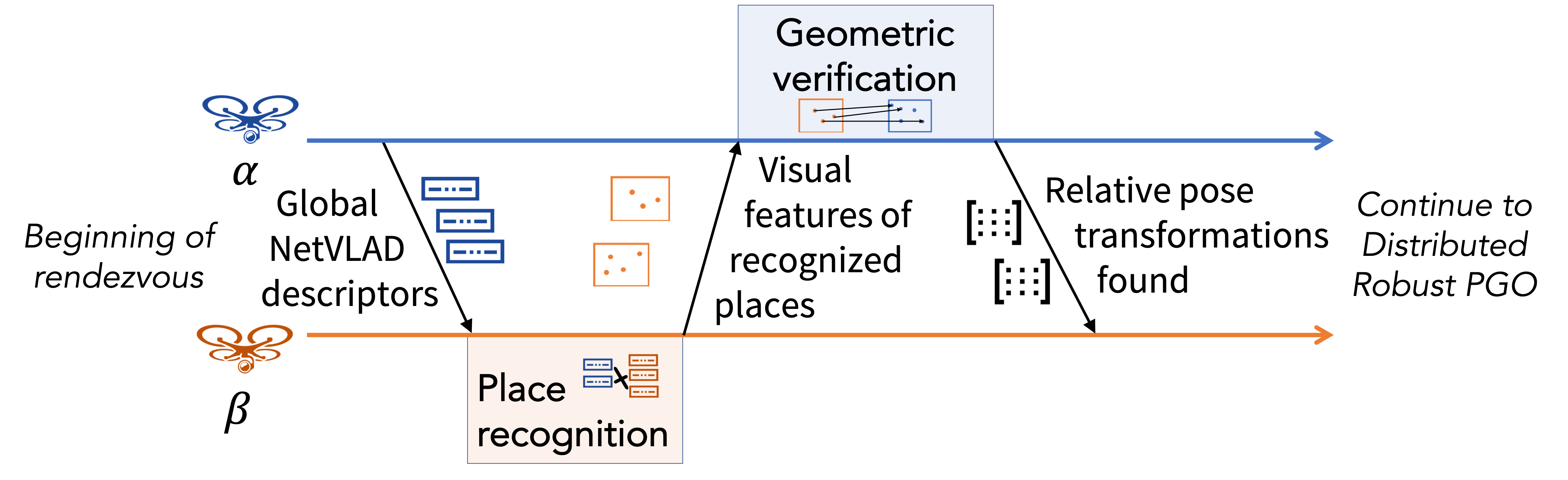} 
\caption{ Distributed loop closures detection overview.}
\label{fig:distLoopClosures}
\vspace{-7mm}
\end{figure}

The distributed loop closure detection includes two submodules.  The first
submodule, \emph{place recognition}, allows to find loop closure candidates
using compact image descriptors.  The second submodule, \emph{geometric
  verification}, computes the relative pose estimate between two robot poses
observing the same scene. The process is illustrated in
Fig.~\ref{fig:distLoopClosures}.

The {\bf place recognition submodule} relies on \NETVLAD
descriptors~\cite{Arandjelovic16cvpr-netvlad} which are compact and robust to
viewpoint and illumination changes.  Each robot locally computes the \NETVLAD
descriptors for each keyframe provided by the stereo visual odometry
module. Once two robots ($\alpha$ and $\beta$) are in communication range, one
of them ($\alpha$) sends \NETVLAD descriptors to the other ($\beta$). Robot
$\alpha$ only sends the descriptors which have been generated since both
robots' last encounter or all of them if it is their first
rendezvous. Robot $\beta$ compares the received \NETVLAD descriptors against
the ones it has generated from its own keyframes. By doing so, robot $\beta$
selects potential loop closures corresponding to pairs of keyframes having
Euclidean distance below a given threshold.
This process provides putative loop closures without requiring the exchange of
raw data, full connectivity maintenance, or additional environment-specific
pre-training.

Each robot also extracts visual features from the left image of the stereo
pair, the associated feature descriptors, and their corresponding estimated 3D
positions; these are used by the {\bf geometric verification submodule}.
\modification{After finding a set of putative loop closures, robot $\beta$ sends the visual features, along with their descriptors and 3D positions, back to robot
$\alpha$. This is done for each keyframe involved in a putative loop closure.}
Using these features, robot $\alpha$ performs geometric verification using the
solvePnpRansac function from \OPENCV~\cite{opencv_library}, which returns a
set of inlier features and a relative pose transformation.
If the set of inliers is sufficiently large (see Section~\ref{sec:experiments}),  
 robot $\alpha$ considers the corresponding loop closure successful.
 Finally, robot $\alpha$ communicates back the relative poses corresponding to
 successful loop closures to robot $\beta$.
 Once the inter-robot loop closures are found and shared, both robots 
 initiate the distributed robust pose graph optimization protocol described in
 the following section.

\subsection{Distributed Robust PGO}
\label{sec:dist_robust_pgo}

This module is in charge of estimating the robots' trajectories given the
odometry measurements from the stereo visual odometry module and the relative
pose measurements from the distributed loop closure detection module.  The
module also includes a distributed outlier rejection approach that removes
spurious loop closures that may accidentally pass the geometric verification
step described in Section~\ref{sec:dist_loop_closure_detection}.

The (to-be-computed) trajectory of each robot is represented as a discrete set
of poses, describing the position and the orientation of its camera at each
keyframe. %
We denote the trajectory of robot $\alpha$ as
$\vxx_\alpha\doteq[\boldsymbol{x}_{\alpha_0}, \boldsymbol{x}_{\alpha_1},
...]$, where $x_{\alpha_i} = [\MR_{\alpha_i}, \vt_{\alpha_i}] \in \SE3$, and
$\MR_{\alpha_i} \in \SO3$ and $\vt_{\alpha_i} \in \Real3$ represent the
rotation and the translation of the pose associated to the $i$-th keyframe of
robot $\alpha$.

The stereo visual odometry module produces odometry measurements,
describing the relative pose between consecutive keyframes: for instance, $\barz_{\alpha_i}^{\alpha_{i-1}} \doteq [\barR_{\alpha_i}^{\alpha_{i-1}}, \bart_{\alpha_i}^{\alpha_{i-1}}]$, denotes the (measured) motion of robot $\alpha$ between keyframe $i-1$ and keyframe $i$. 
On the other hand, the distributed loop closure detection module produces noisy
relative pose measurements of the relative pose of two robots observing the same place: 
for instance, the inter-robot measurement 
$\barz_{\beta_k}^{\alpha_i} \doteq [\barR_{\beta_k}^{\alpha_i}, \bart_{\beta_k}^{\alpha_i}]$ describes a measurement of the relative pose between the $i$-th keyframe of robot $\alpha$ and the $k$-th keyframe of robot $\beta$.

 Our system includes two submodules: distributed outlier rejection and distributed pose graph optimization.
The {\bf distributed outlier  rejection submodule} rejects spurious inter-robot loop closures 
$\barz_{\beta_k}^{\alpha_i}$ that may be caused by perceptual aliasing; 
if undetected, these outliers cause large distortions in the robot trajectory estimates (Fig.~\ref{fig:fieldExperimentsTrajectories}).

\begin{figure}[t!]
\centering
  \includegraphics[width=0.75\linewidth]{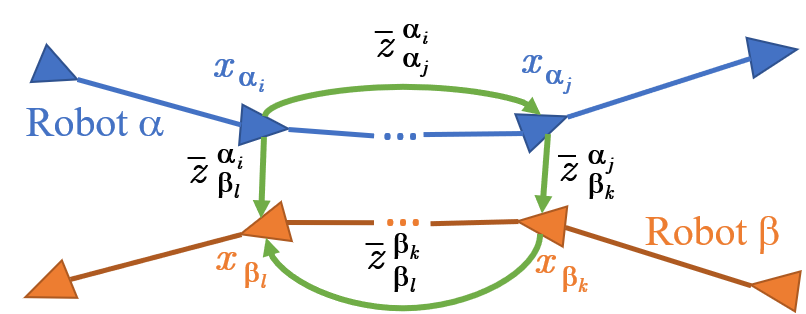}
  \vspace{-2mm}
\caption{Measurements needed to check pairwise consistency.}
\label{fig:PCM}
\vspace{-7mm}
\end{figure}

 We adopt the \emph{Pairwise Consistent Measurement Set Maximization} (\PCM) technique proposed by Mangelson~\etal~\cite{Mangelson18icra} for outlier rejection and 
tailor it to a fully distributed setup. %
The key insight behind \PCM is to check if pairs of inter-robot loop closures
are consistent with each other and then search for a large set of
mutually-consistent loop closures (as shown in~\cite{Mangelson18icra}, the
largest set of pairwise consistent measurements can be found as a maximum
clique).
Although \PCM does not check for the joint consistency of all the
measurements, the approach typically ensures that gross outliers are rejected.
The following metric is used to determine if two inter-robot loop closures
$\barz_{\beta_k}^{\alpha_j}$ and $\barz_{\beta_l}^{\alpha_i}$ are pairwise
consistent:
\beq
\| ( \barz_{\alpha_j}^{\alpha_i} \oplus \barz_{\beta_k}^{\alpha_j} \oplus \barz_{\beta_l}^{\beta_k} ) \ominus \barz_{\beta_l}^{\alpha_i} \|_{\Sigma} \leq \gamma
\label{eq:consistencyMetric}
\eeq
In this equation, $\| \cdot \|_{\Sigma}$ represents the Mahalanobis distance and we use the notation of~\cite{Smith87} to denote the pose composition $\oplus$ and inversion $\ominus$.  
Intuitively, in the noiseless case, measurements along the cycle (shown in green in Fig.~\ref{fig:PCM}) formed by the loop closures 
($\barz_{\beta_l}^{\alpha_i}$, $\barz_{\beta_k}^{\alpha_j}$) and the odometry 
($\barz_{\alpha_j}^{\alpha_i}$, $\barz_{\beta_l}^{\beta_k}$) must compose to the identity, and the consistency metric~\eqref{eq:consistencyMetric} assesses that the noise accumulated along the cycle is consistent with the noise covariance $\Sigma$.
The \PCM likelihood threshold $\gamma$ can be determined from the quantile of the chi-squared distribution 
for a given probability level~\cite{Mangelson18icra}.

The key insight of this section is that the consistency metric~\eqref{eq:consistencyMetric} 
can be computed from the loop closure measurements ($\barz_{\beta_l}^{\alpha_i}$, $\barz_{\beta_k}^{\alpha_j}$) 
and the \modification{odometric estimates of the poses involved} ($x_{\alpha_i}, \modification{x_{\alpha_j}}, x_{\beta_l}, x_{\beta_k} )$. Since both quantities are already used in the distributed PGO algorithm (described below), 
the outlier rejection can be performed ``for free'', without requiring extra communication.
\modification{After the pairwise consistency checks are performed, each robot computes the maximum clique of the measurements for each of its neighbors to find inlier loop closures. The inliers are passed to the distributed PGO.}

The {\bf distributed PGO submodule} uses the odometry measurements and the
inlier inter-robot loop closures to compute the trajectory estimates of the
robots.
 We use the approach proposed in~\cite{Choudhary17ijrr-distributedPGO3D}:
 the robots repeatedly exchange their \modification{estimate for the poses involved in inter-robot loop closures}
 till they reach a consensus on the optimal trajectory estimate.
More specifically, the approach of~\cite{Choudhary17ijrr-distributedPGO3D}
solves pose graph optimization in a distributed fashion using a two-stage
approach: first, it computes an estimate for the rotations of the robots along
their trajectories; and then it recovers the full poses in a second stage.
Each stage can be solved using a distributed Gauss-Seidel
 algorithm~\cite{Choudhary17ijrr-distributedPGO3D} which avoids complex
 bookkeeping and information double counting, and
requires minimal information exchange.

\section{Experimental Results}
\label{sec:experiments}

This section presents four sets of experiments. Section~\ref{sec:simulation}
tests the performance of the outlier rejection mechanism in a simulated
multi-robot SLAM environment. Section~\ref{sec:dataset} evaluates the results
of \DOORSLAM on the widely used \KITTI
sequence~\cite{Geiger12cvpr}. Section~\ref{sec:field1} reports the results of
field experiments conducted with two flying drones on an outdoor football
field. Finally, Section~\ref{sec:field2} reports the results of field tests
conducted in underground environments in the context of the DARPA Subterranean
Challenge~\cite{DARPASubT}.

\subsection{Implementation Details}
\label{sec:implementation}

The \DOORSLAM system is the result of the combination of many frameworks and
libraries. First, we use the Robot Operating System to interface with the
onboard camera and handle information exchange between the different core
modules.  We use the \BUZZ~\cite{Pinciroli2016} programming language and runtime
environment for communication and scheduling.  In the front-end, we use the
latest version of~\RTABMAP~\cite{Labbe2019} for stereo visual odometry and we
use the tensorflow implementation of \NETVLAD provided
in~\cite{Cieslewski18icra}, with the default neural network weights trained in
the original paper~\cite{Arandjelovic16cvpr-netvlad}. We only keep the first
128 dimensions of the generated descriptors to limit the data to be exchanged,
as done in~\cite{Cieslewski18icra}. The visual feature extraction and relative
pose transformation estimation are done by adapting the implementation in
\RTABMAP and keeping their default parameters. The features used are Good
Features to Track~\cite{Shi94} with ORB descriptors~\cite{rublee2011iccv-orb}.
We implemented the distributed robust PGO module in C++ using the GTSAM
library~\cite{Dellaert12tr} and building on the implementation of Choudhary et
al.~\cite{Choudhary17ijrr-distributedPGO3D}.  We followed a simulation,
software-in-the-loop, hardware-in-the-loop, robot deployment code base
implementation paradigm, starting from \ARGOS simulation and ending with full
deployment using {\smaller\sf Docker} containers on {\smaller\sf NVIDIA Jetson
  TX2} on-board computers.

\subsection{Simulation Experiments} 
\label{sec:simulation}

To verify that our online and distributed implementation of \PCM is able to
correctly reject outliers, we designed a simulation using \ARGOS~\cite{Pinciroli2012}.  We refer
the reader to the video attachment for a visualization.  We use 5
drones with limited communication range following random trajectories. We
simulate the SLAM front-end by building their respective pose graphs using
noisy measurements. When two robots come within communication range,
they exchange inter-robot measurements based on their current poses and then
use our SLAM back-end (\PCM + distributed PGO) to compute a shared pose graph
solution in a fully distributed manner. Inlier inter-robot loop closures are
added with realistic Gaussian noise ($\sigma_{R}=0.01$rad and $\sigma_{t}=0.1$m
for rotation and translation measurements, respectively) while outliers are
sampled from a uniform distribution.

\begin{figure}[h]
\vspace{-3mm}
	\centering	
	\begin{minipage}{\columnwidth}
		\centering
			\includegraphics[width=\textwidth,trim=0mm 0mm 0mm 0mm,clip]{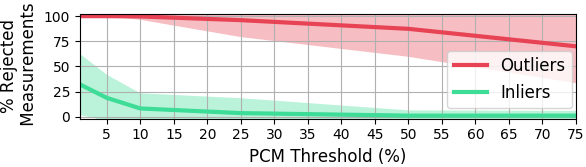}
	\end{minipage}
	\vspace{-3mm}
	\caption{\modification{Percentage of inliers and outliers rejected w.r.t. \PCM likelihood threshold (100 runs avg. $\pm$ std.) in \ARGOS.} }
	\label{fig:simulationPCM}
\vspace{-2mm}
\end{figure}

\begin{figure}[h]
\vspace{-3mm}
	\centering	
	\begin{minipage}{\columnwidth}
		\centering
			\includegraphics[width=\textwidth,trim=0mm 0mm 0mm 0mm,clip]{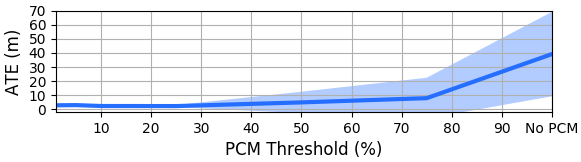}
	\end{minipage}
	\vspace{-3mm}
	\caption{\modification{Average Translation Error (ATE) w.r.t. \PCM likelihood threshold (10 runs avg. $\pm$ std.) in \ARGOS.}} 

	\label{fig:simulationPCM_ATE}
\vspace{-2mm}
\end{figure}

\textbf{Results.} %
\modification{We look at three metrics in particular: 
the percentage of outliers rejected,
the percentage of inliers rejected and the average translation error (ATE). The first evaluates if the spurious
measurements are successfully rejected; 
the ideal value for this metric is 100\%. The second indicates if the
technique is needlessly rejecting valid measurements; the ideal value is
0\%.} \modification{The third evaluates the distortion of the estimates. Fig.~\ref{fig:simulationPCM} shows the percentage of outliers (in red) and
inliers (in green) rejected with different \PCM
thresholds while Fig.~\ref{fig:simulationPCM_ATE} shows the ATE (in blue)}; the threshold represents the likelihood of accepting an outlier as
inlier. \modification{As expected, using a lower threshold leads to the rejection of more
measurements, including inliers, while using a higher threshold can lead to
the occasional acceptance of outliers which in turn leads to a larger error.} %
Therefore, in all our
experiments, we used a threshold of 1\% to showcase the performance of our
system in its safest configuration.

\subsection{Dataset Experiments} 
\label{sec:dataset}

\begin{figure}[t]
\centering
\begin{subfigure}[b]{0.23\textwidth}
\centering
\includegraphics[width=\textwidth,trim=0mm 0mm 0mm 0mm,clip]{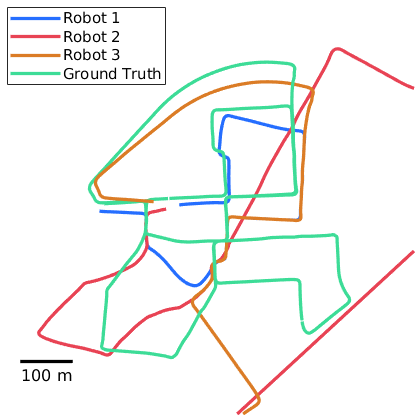}
\caption{\modification{\textbf{without} outlier rejection.}}
\end{subfigure}
\begin{subfigure}[b]{0.23\textwidth}
\centering
\includegraphics[width=\textwidth,trim=0mm 0mm 0mm 0mm,clip]{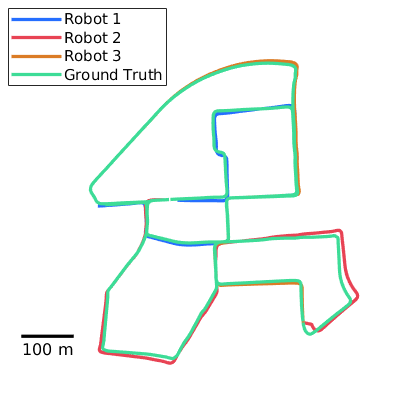}\caption{\modification{\textbf{with} outlier rejection.}}
\end{subfigure}
\vspace{-2mm}
\caption{\modification{Experiment on the \KITTI dataset. Optimized trajectories (red, blue, and orange) and ground truth (green).}}
\label{fig:KittiTrajectories}
\vspace{-1mm}
\end{figure}

The \KITTI~\cite{Geiger12cvpr} sequence is a popular benchmark for SLAM. \modification{In
our evaluation, we split the sequence into three parts and execute \DOORSLAM on
three {\smaller\sf NVIDIA Jetson TX2}s.} We used a \PCM threshold of 1\%, a
\NETVLAD comparison threshold of $0.15$, and a minimum of $5$ feature
correspondences in the geometric verification to get a high number of loop closure measurements. 
While related work uses more
conservative thresholds for \NETVLAD and the number of feature correspondences to avoid
outliers~\cite{Cieslewski18icra}, we can afford more aggressive thresholds
thanks to \PCM.

\textbf{Results.} Fig.~\ref{fig:KittiTrajectories} shows that outliers are
present among the loop closure measurements and that their effect on the pose
graph is significant. \modification{The average translation error (ATE) without outlier
rejection is 86.85m, while the error is reduced to 8.00m when using \PCM.}
It is important to note that the error is higher than recent SLAM solutions on
this sequence since for simplicity's sake we do not make use of any
\emph{intra-robot} loop closures. \modification{Additionnal results on other \KITTIs sequences are available in the supplemental material~\cite{Lajoie19tr-DOORSLAM}}.

\subsection{Field Tests with Drones} 
\label{sec:field1}

\begin{figure}[t]
\hspace{3mm}
\begin{minipage}{0.33\columnwidth} 
\centering
\vspace{-3mm}
\includegraphics[width=\textwidth,trim=0mm 0mm 0mm 0mm,clip]{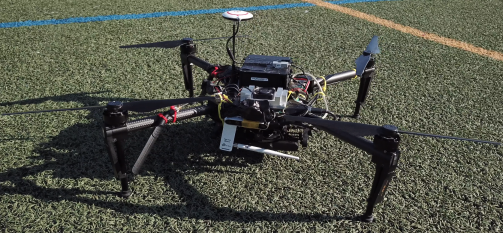}\vspace{-3mm}
\end{minipage}%
\begin{minipage}{0.67\columnwidth} 
\begin{tabular}{|l||l|}
	\hline
	Platform & \smaller{DJI Matrice 100} \\
	\hline
    Camera & \smaller{Intel Realsense D435} \\
	\hline
    Computer & \smaller{NVIDIA Jetson TX2} \\
	\hline
\end{tabular}
\end{minipage}
\vspace{-2mm}
\caption{Hardware setup used in field experiments.}
\label{fig:hardware}
\vspace{-6mm}
\end{figure}

To test that \DOORSLAM can overcome the reality gap and map environments with
severe perceptual aliasing using
resource-constrained platforms, we also performed field experiments with two
quadcopters featuring stereo cameras, flying over a football field.
The cameras facing slightly downward are subject to perceptual aliasing, due
to the repetitive appearance of the field (see video attachment).
The hardware setup is described in Fig.~\ref{fig:hardware}.

We performed manual flights with trajectories approximately following simple
geometric shapes as seen in Fig.~\ref{fig:fieldExperimentsTrajectories}. For
the first experiments we recorded images and GPS data on the field and we
executed \DOORSLAM in an offline fashion on two {\smaller\sf NVIDIA Jetson
  TX2} connected through WiFi.  This allowed us to reuse the same recordings
with various combinations of the three major parameters of \DOORSLAM and study
their influence (Section~\ref{subsubsec:param_influence}) as well as assess \DOORSLAM's communication requirements 
(Section~\ref{subsubsec:communication}).
Finally, we performed an online experiment where \DOORSLAM is executed on the
drones' onboard computers during flight (see Section
\ref{subsubsec:onlineexperiments} and video attachment).

\subsubsection{Influence of Parameters} \label{subsubsec:param_influence}
As practitioners know, SLAM systems often rely on precise parameter tuning,
especially to avoid outlier measurements from the front-end. We show that
\DOORSLAM is less sensitive to the parameter tuning since our back-end can
handle spurious measurements. Moreover, we can leverage the robustness to
outliers to significantly increase the number of loop closure candidates and
potentially the number of valid measurements.

\begin{figure}
	\begin{minipage}{\columnwidth} 
	\centering
	\includegraphics[width=\textwidth,trim=0mm 0mm 0mm 0mm,clip]{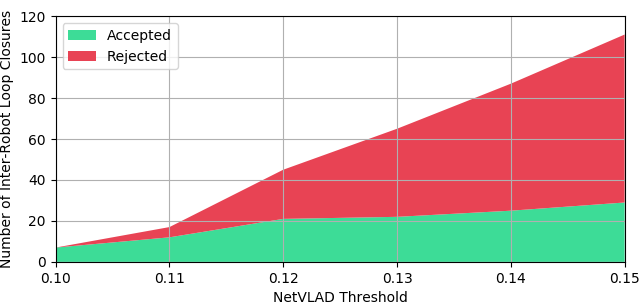}
	\end{minipage}
\vspace{-2mm}
	\caption{Number of inter-robot loop closures accepted and rejected by \PCM w.r.t. the \NETVLAD threshold. We fix the minimum number of feature correspondences to 5.}
	\label{fig:effectNetVLADThreshold}
\vspace{-2mm}
\end{figure}

\textbf{Results.} In many scenarios, loop closures are hard to obtain due to
external conditions such as illumination changes. Hence, it is important to
consider as many loop closure candidates as possible. Instead of rejecting
them prematurely in the front-end, \DOORSLAM can consider more candidates and
only reject the outliers
before the optimization. To analyze the gain of being less conservative, we
looked at the number of inter-robot loop closures detected with various
\NETVLAD thresholds (Fig.~\ref{fig:effectNetVLADThreshold}). As expected, when
we increase this threshold, we obtain more candidates. Interestingly, even
though most of the new loop closures are rejected by \PCM (in red), we also
get about three times more valid measurements (green) when using a looser
threshold (0.15) as opposed to a more conservative one (0.10). Therefore, the
use of less stringent thresholds allows adding valid measurements to the pose
graph, enhancing the trajectory estimation accuracy.
\begin{figure}
	\begin{minipage}{\columnwidth} 
	\centering
	\includegraphics[width=\textwidth,trim=0mm 0mm 0mm 0mm,clip]{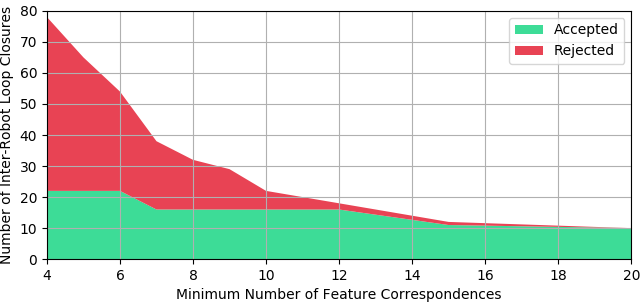}
	\end{minipage}
\vspace{-2mm}
	\caption{Number of inter-robot loop closures accepted and rejected by \PCM w.r.t. the minimum number of feature correspondences to consider geometric verification successful. We fix the \NETVLAD threshold to 0.13
	}
	\label{fig:effectMinInliers}
\vspace{-2mm}
\end{figure}

Similarly, reducing the minimum number of feature correspondences that need to
pass the geometric verification step for a loop closure to be considered
successful
leads to more loop closure candidates. \RTABMAP uses a default of 20
correspondences.
As shown in Fig.~\ref{fig:effectMinInliers}, we can double the number of valid
inter-robot loop closures when reducing the number of correspondences to 4 or
5.

\begin{table}
  \centering
  \setlength\tabcolsep{4pt}
  \caption{Effect of the \PCM threshold on the accuracy.} 
  \vspace{-2mm}
  \begin{tabular}{|c||c|c|c|c|c|c|}
    \hline
    Threshold (\%)& 1 & 10 & 25 & 75 & No \PCM \\
    \hhline{=======}
    ATE (m)& 2.1930 & 2.3185 & 3.1461 & 18.255 & 22.0159 \\
    \hline
  \end{tabular}
  \label{tab:fieldExperimentsErrors}
   \vspace{-6mm}
\end{table}

The last parameter we analyzed is the \PCM likelihood threshold to reject
outliers. As seen in Section~\ref{sec:simulation}, a lower threshold leads to
the rejection of more measurements, including inliers. However, since we are
mapping a relatively small environment, we get many loop closures linking the
same places. Therefore, as long as we do not disconnect the recognized places
in the pose graph, a lower \PCM threshold has the benefit of filtering out the
noisiest loop closures and keeping the more precise ones.  We can see in Table
\ref{tab:fieldExperimentsErrors} that the resulting trajectories are affected
by the noisier loop closures when we use a higher threshold, but that we still
avoid the dramatic distortion caused by outliers seen in
Fig.~\ref{fig:fieldExperimentsTrajectories}.  Indeed, the average translation
error (ATE) compared to the GPS ground truth is the lowest when we use the most
conservative \PCM threshold (i.e. 1\%), for which we show the visual result in
Fig.~\ref{fig:fieldExperimentsTrajectories}. On the other hand, we can see a
large increase in the error when we use a threshold larger than 75\% or no
\PCM, which indicates that outliers have not been rejected.

In light of those results, \DOORSLAM can use less conservative parameters in
the front-end to obtain more loop closure candidates and a more conservative
\PCM threshold to keep only the most accurate ones. This combination leads to a
larger number valid loop closures and to more accurate trajectory estimates.

\subsubsection{Communication}\label{subsubsec:communication}

\begin{table}[]
\caption{Data sizes of messages sent.}
\vspace{-2mm}
\label{tab:data_sizes}
\centering
\resizebox{0.49\textwidth}{!}{%
\begin{tabular}{|c|l|c|}
\hline
\multicolumn{2}{|c|}{\textbf{Details of message sent for each}} & Avg. Size (kB) $\pm$ Std.
\\ \hhline{===}
\multirow{2}{0.3\columnwidth}{\textbf{Keyframe}} & {\NETVLAD descriptor} & 1.00  $\pm$  0.00 
\\ \hhline{~--}  
& \cellcolor[gray]{0.9}\textit{RGB image} & \cellcolor[gray]{0.9}\textit{900.04  $\pm$  0.00}
\\ \hhline{===}
\multirow{3}{0.3\columnwidth}{\textbf{\NETVLAD match}}     
& Keypoints Information & 34.51 $\pm$ 0.68 
\\ \hhline{~--}
& Keypoints Descriptors & 25.00 $\pm$ 0.49                       
\\ \hhline{~--} 
&\cellcolor[gray]{0.9}\textit{Grayscale images} & \cellcolor[gray]{0.9}\textit{600.06 $\pm$  0.0} 
\\ \hhline{===}
\multirow{2}{0.3\columnwidth}{\textbf{Inter-robot loop closure}} 
& Pose Estimate & 0.34 $\pm$  0.00
\\ \hhline{~--} 
& Loop Closure Measurement & 0.34   $\pm$  0.00 
\\ \hline
\end{tabular}}
\end{table}

As described in Section \ref{sec:dist_loop_closure_detection}, the distributed
loop closure detection module needs to share information between the robots about each 
keyframe to detect loop closure candidates. When a \NETVLAD match
occurs, the module needs to send the keypoint information for each matching
keyframe. If there are enough feature correspondences, the module can compute
the relative pose transformation and send the resulting inter-robot
measurement to the other robot.  Here we evaluate the communication cost of
the proposed distributed front-end.

\textbf{Results.} Table \ref{tab:data_sizes} reports the average data size
sent at each keyframe. %
These averages were computed during our field experiments.  For comparison, we
also report (in gray) the size of the messages sent in case the
robots were to directly transmit camera images.  We see that the proposed
front-end reduces the required bandwidth by roughly a factor of 10. %

\subsubsection{Online Experiments}\label{subsubsec:onlineexperiments}

We tested \DOORSLAM online with two quadcopters. The main challenge of
performing live experiments with \DOORSLAM on the {\smaller\sf NVIDIA Jetson
  TX2} platforms is to run every module in real-time with the additional
workload of the camera driver and the connection to the flight controller. To
achieve this feat, we limited the frame rate of the onboard camera to
6Hz. Modules such as the stereo odometry or the Tensorflow implementation of
\NETVLAD were particularly demanding in terms of RAM which required us to add
4GB of swap space to the 8GB initially available.  We also tuned some visual
odometry parameters to gain computational performance at the cost of losing some
accuracy.

\begin{figure}[t]
\centering
\begin{subfigure}[b]{0.23\textwidth}
\centering
\includegraphics[width=\textwidth,trim=0mm 0mm 0mm 0mm,clip]{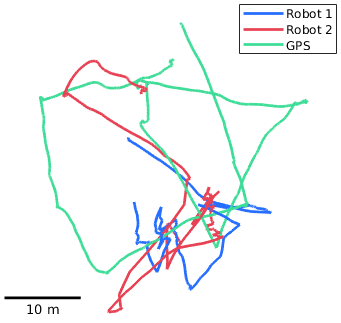}
\caption{\textbf{without} outlier rejection.}
\end{subfigure}
\begin{subfigure}[b]{0.23\textwidth}
\centering
\includegraphics[width=\textwidth,trim=0mm 0mm 0mm 0mm,clip]{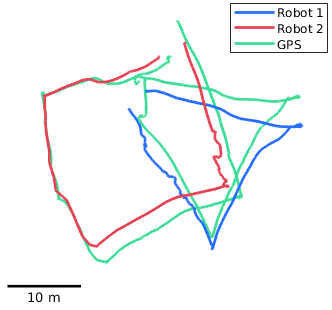}\caption{\textbf{with} outlier rejection.}
\end{subfigure}
\vspace{-2mm}
\caption{Online Trajectory estimates from \DOORSLAM (red and blue) and GPS ground truth (green, only used for benchmarking).}
\label{fig:fieldExperimentsTrajectoriesOnline}
\vspace{-6mm}
\end{figure}

\textbf{Results.} Fig.~\ref{fig:fieldExperimentsTrajectoriesOnline} reports
the trajectory estimates of our online experiments, compared with the
trajectories from GPS. We performed this experiment with a \PCM threshold of
1\%, a \NETVLAD threshold of 0.13, and a minimum of 5 inliers for geometric
verification. Although we note a degradation of the visual odometry accuracy,
the results in Fig.~\ref{fig:fieldExperimentsTrajectoriesOnline} are
consistent with the ones observed in
Fig.~\ref{fig:fieldExperimentsTrajectories}.

\subsection{Field Tests in Subterranean Environments} 
\label{sec:field2}

\begin{figure}[t]
\centering
\begin{subfigure}[b]{0.23\textwidth}
\centering
\includegraphics[width=\textwidth,trim=0mm 0mm 0mm 0mm,clip]{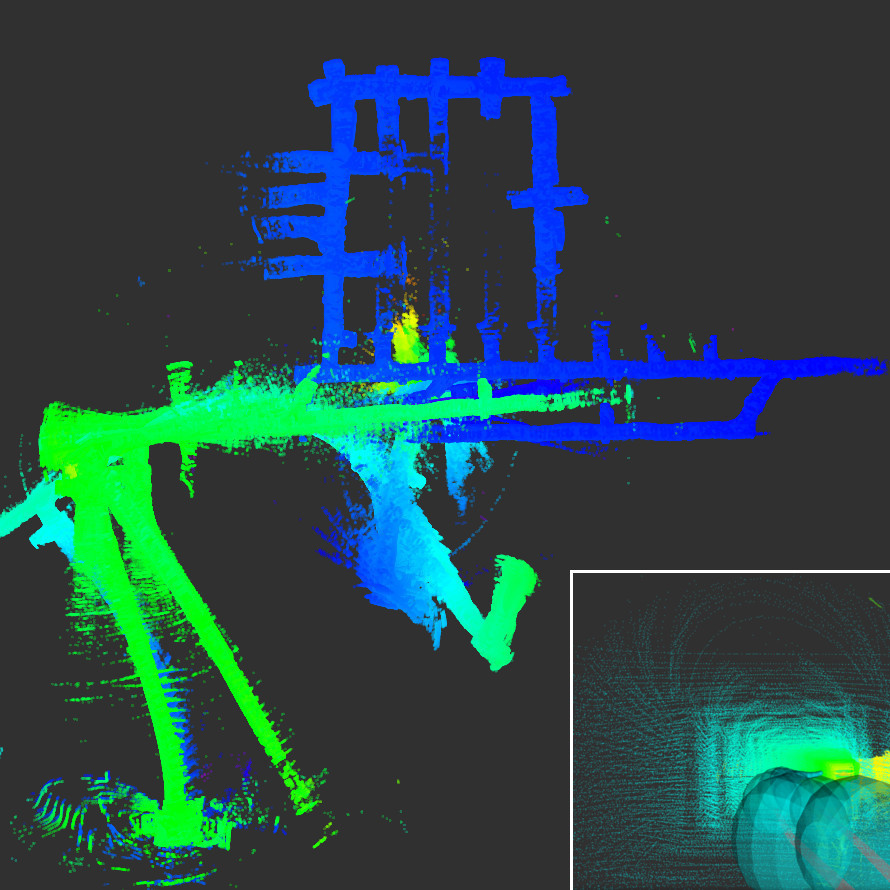}
\caption{\textbf{without} outlier rejection.}
\end{subfigure}
\begin{subfigure}[b]{0.23\textwidth}
\centering
\includegraphics[width=\textwidth,trim=0mm 0mm 0mm 0mm,clip]{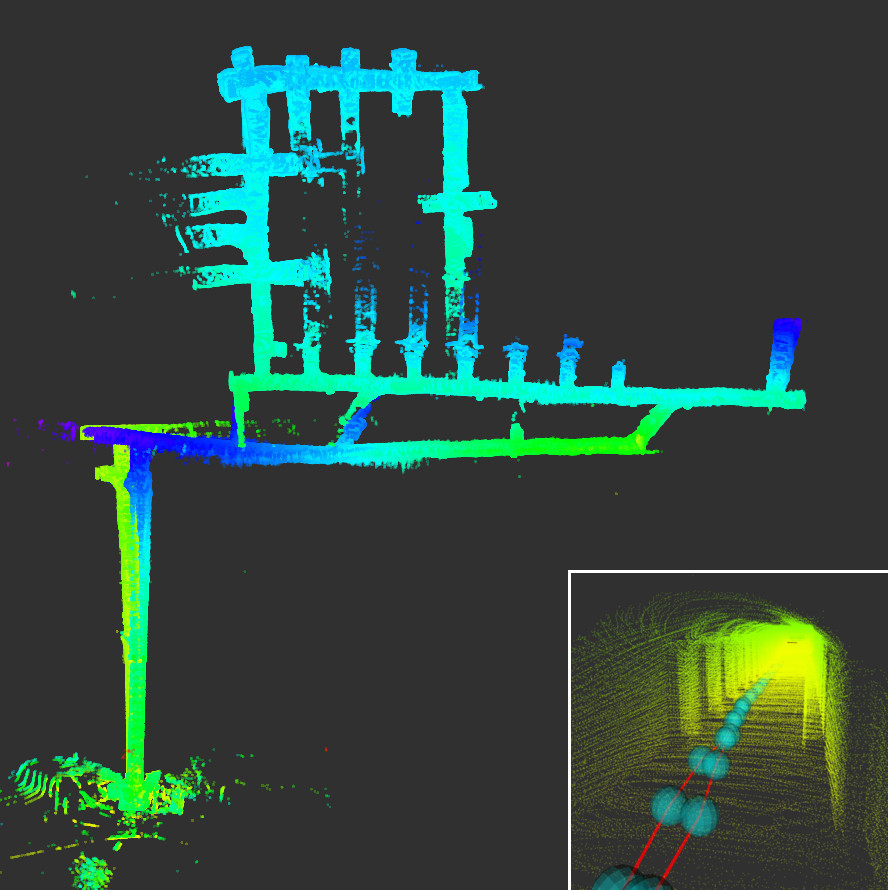}\caption{\textbf{with} outlier rejection.}
\end{subfigure}
\vspace{-2mm}
\caption{Lidar-based multi-robot SLAM experiment during the DARPA Subterranean Challenge.}
\label{fig:DARPA}
\vspace{-6mm}
\end{figure}

To remark on the generality of the \DOORSLAM back-end, this section considers
a different sensor front-end and shows that \DOORSLAM can be used in a
lidar-based SLAM setup with minimal modifications.
For this purpose we used lidar data collected by two 
Husky UGVs during the Tunnel Circuit competition of the DARPA Subterranean
Challenge~\cite{DARPASubT}.
The data is collected with the VLP-16 Puck LITE 3D lidar and the loop closures
are detected by scan matching using ICP. The environment, over 1 kilometer
long, is a coal mine whose self-similar appearance is prone to causing
perceptual aliasing and outliers.
Fig.~\ref{fig:DARPA} shows the effect of using \PCM: the left figure shows a
top-view of the point cloud resulting from multi-robot SLAM without \PCM, while the figure on the
right is produced using \PCM with a threshold of 1\%.  The reader may notice
the deformation on the left figure, caused by an incorrect loop closure
between two different segments of the tunnel.
Although \PCM largely improves the mapping performance, we notice that there is still an incorrect loop closure on the right figure.
This kind of error is likely due to the fact that \PCM requires a correct
estimate of the measurement covariances which is not always available.
To compute the trajectory estimates, our distributed back-end required the
transmission of 92.27kB, while in a centralized setup the transmission of the
initial pose graph data and the resulting estimates from one robot to the
other would require 196.30kB. In summary, our distributed back-end implementation
roughly halves the communication burden.

\section{Conclusion}
\label{sec:conclusion}

We present \DOORSLAM, a system for distributed multi-robot SLAM consisting of a data-efficient peer-to-peer front-end and an outlier-resilient back-end. %
Our experiments in simulation, datasets, and field tests show that our approach rejects spurious measurements and computes accurate trajectory estimates. We also show that our approach can leverage its robust back-end to work with less conservative front-end parameters. 
\modification{In future work, we plan to explore not only the robustness to
  additional perception failures, such as large groups of correlated outliers,
  but also the robustness to communication issues (i.e., packet drop) to
  improve the safety and resilience of multi-robot SLAM systems.}

\bibliographystyle{IEEEtran}

\begin{thebibliography}{10}
\providecommand{\url}[1]{#1}
\csname url@rmstyle\endcsname
\providecommand{\newblock}{\relax}
\providecommand{\bibinfo}[2]{#2}
\providecommand\BIBentrySTDinterwordspacing{\spaceskip=0pt\relax}
\providecommand\BIBentryALTinterwordstretchfactor{4}
\providecommand\BIBentryALTinterwordspacing{\spaceskip=\fontdimen2\font plus
\BIBentryALTinterwordstretchfactor\fontdimen3\font minus
  \fontdimen4\font\relax}
\providecommand\BIBforeignlanguage[2]{{%
\expandafter\ifx\csname l@#1\endcsname\relax
\typeout{** WARNING: IEEEtran.bst: No hyphenation pattern has been}%
\typeout{** loaded for the language `#1'. Using the pattern for}%
\typeout{** the default language instead.}%
\else
\language=\csname l@#1\endcsname
\fi
#2}}

\bibitem{Mangelson18icra}
J.~G. Mangelson, D.~Dominic, R.~M. Eustice, and R.~Vasudevan, ``Pairwise
  consistent measurement set maximization for robust multi-robot map merging,''
  in \emph{IEEE Intl. Conf. on Robotics and Automation (ICRA)}, 2018, pp.
  2916--2923.

\bibitem{cieslewski_efficient_2017-bow}
T.~Cieslewski and D.~Scaramuzza, ``\BIBforeignlanguage{en}{Efficient
  {Decentralized} {Visual} {Place} {Recognition} {Using} a {Distributed}
  {Inverted} {Index}},'' \emph{\BIBforeignlanguage{en}{IEEE Robotics and
  Automation Letters}}, vol.~2, no.~2, pp. 640--647, Apr. 2017.

\bibitem{Choudhary17ijrr-distributedPGO3D}
S.~Choudhary, L.~Carlone, C.~Nieto, J.~Rogers, H.~Christensen, and F.~Dellaert,
  ``Distributed mapping with privacy and communication constraints: Lightweight
  algorithms and object-based models, accepted,'' \emph{Intl. J. of Robotics
  Research}, 2017, arxiv preprint: 1702.03435.

\bibitem{Sunderhauf12iros}
N.~S\"{u}nderhauf and P.~Protzel, ``Switchable constraints for robust pose
  graph {SLAM},'' in \emph{IEEE/RSJ Intl. Conf. on Intelligent Robots and
  Systems (IROS)}, 2012.

\bibitem{Agarwal13icra}
P.~Agarwal, G.~D. Tipaldi, L.~Spinello, C.~Stachniss, and W.~Burgard, ``Robust
  map optimization using dynamic covariance scaling,'' in \emph{IEEE Intl.
  Conf. on Robotics and Automation (ICRA)}, 2013.

\bibitem{Olson12rss}
E.~Olson and P.~Agarwal, ``Inference on networks of mixtures for robust robot
  mapping,'' in \emph{Robotics: Science and Systems (RSS)}, July 2012.

\bibitem{Latif14rss}
L.~Yasir, G.~Huang, J.~Leonard, and J.~Neira, ``{An Online Sparsity-cognizant
  Algorithm for Visual Navigation},'' in \emph{Robotics: Science and Systems
  (RSS)}, 2014.

\bibitem{Lajoie19ral-DCGM}
P.~Lajoie, S.~Hu, G.~Beltrame, and L.~Carlone, ``Modeling perceptual aliasing
  in {SLAM} via discrete-continuous graphical models,'' \emph{{IEEE} Robotics
  and Automation Letters ({RA-L})}, 2019, extended ArXiv version:
  \linkToPdf{https://arxiv.org/pdf/1810.11692.pdf}, Supplemental Material:
  \linkToPdf{https://www.dropbox.com/s/vupak65wi75yzbl/2018j-RAL-DCGM-supplemental.pdf?dl=0}.

\bibitem{Cieslewski18icra}
T.~Cieslewski, S.~Choudhary, and D.~Scaramuzza, ``Data-efficient decentralized
  visual {SLAM},'' \emph{IEEE Intl. Conf. on Robotics and Automation (ICRA)},
  2018.

\bibitem{Arandjelovic16cvpr-netvlad}
R.~{Arandjelovic}, P.~{Gronat}, A.~{Torii}, T.~{Pajdla}, and J.~{Sivic},
  ``{NetVLAD}: {CNN} architecture for weakly supervised place recognition,'' in
  \emph{IEEE Conf. on Computer Vision and Pattern Recognition (CVPR)}, 2016,
  pp. 5297--5307.

\bibitem{Geiger12cvpr}
A.~Geiger, P.~Lenz, and R.~Urtasun, ``Are we ready for autonomous driving? the
  {KITTI} vision benchmark suite,'' in \emph{IEEE Conf. on Computer Vision and
  Pattern Recognition (CVPR)}, Providence, USA, June 2012, pp. 3354--3361.

\bibitem{Andersson08icra}
L.~Andersson and J.~Nygards, ``C-{SAM} : Multi-robot {SLAM} using square root
  information smoothing,'' in \emph{IEEE Intl. Conf. on Robotics and Automation
  (ICRA)}, 2008.

\bibitem{Kim10icra}
B.~Kim, M.~Kaess, L.~Fletcher, J.~Leonard, A.~Bachrach, N.~Roy, and S.~Teller,
  ``Multiple relative pose graphs for robust cooperative mapping,'' in
  \emph{IEEE Intl. Conf. on Robotics and Automation (ICRA)}, Anchorage, Alaska,
  May 2010, pp. 3185--3192.

\bibitem{Bailey11icra}
T.~Bailey, M.~Bryson, H.~Mu, J.~Vial, L.~McCalman, and H.~Durrant-Whyte,
  ``Decentralised cooperative localisation for heterogeneous teams of mobile
  robots,'' in \emph{IEEE Intl. Conf. on Robotics and Automation (ICRA)},
  Shanghai, China, May 2011, pp. 2859--2865.

\bibitem{Lazaro11icra}
M.~Lazaro, L.~Paz, P.~Pinies, J.~Castellanos, and G.~Grisetti, ``Multi-robot
  {SLAM} using condensed measurements,'' in \emph{IEEE Intl. Conf. on Robotics
  and Automation (ICRA)}, 2011, pp. 1069--1076.

\bibitem{Dong15icra}
J.~Dong, E.~Nelson, V.~Indelman, N.~Michael, and F.~Dellaert, ``Distributed
  real-time cooperative localization and mapping using an uncertainty-aware
  expectation maximization approach,'' in \emph{IEEE Intl. Conf. on Robotics
  and Automation (ICRA)}, Seattle, WA, May 2015, pp. 5807--5814.

\bibitem{Aragues11icra-distributedLocalization}
R.~Aragues, L.~Carlone, G.~Calafiore, and C.~Sagues, ``Multi-agent localization
  from noisy relative pose measurements,'' in \emph{IEEE Intl. Conf. on
  Robotics and Automation (ICRA)}, 2011, pp. 364--369.

\bibitem{Cunningham10iros}
A.~Cunningham, M.~Paluri, and F.~Dellaert, ``{DDF-SAM}: Fully distributed slam
  using constrained factor graphs,'' in \emph{IEEE/RSJ Intl. Conf. on
  Intelligent Robots and Systems (IROS)}, 2010.

\bibitem{Cunningham13icra}
A.~Cunningham, V.~Indelman, and F.~Dellaert, ``{DDF-SAM} 2.0: Consistent
  distributed smoothing and mapping,'' in \emph{IEEE Intl. Conf. on Robotics
  and Automation (ICRA)}, Karlsruhe, Germany, May 2013.

\bibitem{Wang19arxiv}
W.~Wang, N.~Jadhav, P.~Vohs, N.~Hughes, M.~Mazumder, and S.~Gil, ``Active
  rendezvous for multi-robot pose graph optimization using sensing over
  {Wi-Fi},'' \emph{CoRR}, vol. abs/1907.05538, 2019.

\bibitem{Fischler81}
M.~Fischler and R.~Bolles, ``Random sample consensus: a paradigm for model
  fitting with application to image analysis and automated cartography,''
  \emph{Commun. ACM}, vol.~24, pp. 381--395, 1981.

\bibitem{Neira01tra}
J.~Neira and J.~Tard{\'o}s, ``Data association in stochastic mapping using the
  joint compatibility test,'' \emph{{IEEE} Trans. Robot. Automat.}, vol.~17,
  no.~6, pp. 890--897, December 2001.

\bibitem{Bosse17fnt}
M.~Bosse, G.~Agamennoni, and I.~Gilitschenski, ``Robust estimation and
  applications in robotics,'' \emph{Foundations and Trends in Robotics},
  vol.~4, no.~4, pp. 225--269, 2016.

\bibitem{Hartley13ijcv}
R.~Hartley, J.~Trumpf, Y.~Dai, and H.~Li, ``Rotation averaging,'' \emph{IJCV},
  vol. 103, no.~3, pp. 267--305, 2013.

\bibitem{Pfingsthorn13ijrr}
M.~Pfingsthorn and A.~Birk, ``Simultaneous localization and mapping with
  multimodal probability distributions,'' \emph{Intl. J. of Robotics Research},
  vol.~32, no.~2, pp. 143--171, 2013.

\bibitem{Pfingsthorn16ijrr}
------, ``Generalized graph {SLAM}: Solving local and global ambiguities
  through multimodal and hyperedge constraints,'' \emph{Intl. J. of Robotics
  Research}, vol.~35, no.~6, pp. 601--630, 2016.

\bibitem{Carlone18ral-robustPGO2D}
L.~Carlone and G.~Calafiore, ``Convex relaxations for pose graph optimization
  with outliers,'' \emph{{IEEE} Robotics and Automation Letters ({RA-L})},
  vol.~3, no.~2, pp. 1160--1167, 2018, arxiv preprint: 1801.02112,
  \linkToPdf{https://arxiv.org/pdf/1801.02112.pdf}.

\bibitem{Carlone14iros-robustPGO2D}
L.~Carlone, A.~Censi, and F.~Dellaert, ``Selecting good measurements via
  $\ell_1$ relaxation: a convex approach for robust estimation over graphs,''
  in \emph{IEEE/RSJ Intl. Conf. on Intelligent Robots and Systems (IROS)},
  2014,
  \linkToPdf{https://www.dropbox.com/s/7f304d5ag245ie4/2014c-IROS-outlierRejection.pdf?dl=0}.

\bibitem{Graham15iros}
M.~Graham, J.~How, and D.~Gustafson, ``Robust incremental {SLAM} with
  consistency-checking,'' in \emph{IEEE/RSJ Intl. Conf. on Intelligent Robots
  and Systems (IROS)}, Sept 2015, pp. 117--124.

\bibitem{Oliva01ijcv}
A.~Oliva and A.~Torralba, ``Modeling the shape of the scene: a holistic
  representation of the spatial envelope,'' \emph{Intl. J. of Computer Vision},
  vol.~42, pp. 145--175, 2001.

\bibitem{Ulrich00icra}
I.~Ulrich and I.~Nourbakhsh, ``Appearance-based place recognition for
  topological localization,'' in \emph{IEEE Intl. Conf. on Robotics and
  Automation (ICRA)}, vol.~2, April 2000, pp. 1023 -- 1029.

\bibitem{Lowe99iccv}
D.~Lowe, ``Object recognition from local scale-invariant features,'' in
  \emph{Intl. Conf. on Computer Vision (ICCV)}, 1999, pp. 1150--1157.

\bibitem{Bay06eccv}
H.~Bay, T.~Tuytelaars, and L.~V. Gool, ``Surf: speeded up robust features,'' in
  \emph{European Conf. on Computer Vision (ECCV)}, 2006.

\bibitem{Sivic03iccv}
J.~Sivic and A.~Zisserman, ``Video google: a text re- trieval approach to
  object matching in videos,'' in \emph{Intl. Conf. on Computer Vision (ICCV)},
  2003.

\bibitem{suenderhauf_performance_2015}
N.~Suenderhauf, S.~Shirazi, F.~Dayoub, B.~Upcroft, and M.~Milford, ``On the
  performance of {ConvNet} features for place recognition,'' in \emph{2015
  {IEEE}/{RSJ} {International} {Conference} on {Intelligent} {Robots} and
  {Systems} ({IROS})}, Sept. 2015, pp. 4297--4304.

\bibitem{philbin_object_2007}
J.~Philbin, O.~Chum, M.~Isard, J.~Sivic, and A.~Zisserman, ``Object retrieval
  with large vocabularies and fast spatial matching,'' in \emph{2007 {IEEE}
  {Conference} on {Computer} {Vision} and {Pattern} {Recognition}}, June 2007,
  pp. 1--8.

\bibitem{Scaramuzza11ram}
D.~Scaramuzza and F.~Fraundorfer, ``Visual odometry: Part {I} the first 30
  years and fundamentals,'' 2011.

\bibitem{Tardioli2015}
D.~{Tardioli}, E.~{Montijano}, and A.~R. {Mosteo}, ``Visual data association in
  narrow-bandwidth networks,'' in \emph{IEEE/RSJ International Conference on
  Intelligent Robots and Systems (IROS)}, Sep. 2015, pp. 2572--2577.

\bibitem{cieslewski_efficient_2017-netvlad}
T.~{Cieslewski} and D.~{Scaramuzza}, ``Efficient decentralized visual place
  recognition from full-image descriptors,'' in \emph{2017 International
  Symposium on Multi-Robot and Multi-Agent Systems (MRS)}, Dec 2017, pp.
  78--82.

\bibitem{tian_near-optimal_2018}
Y.~Tian, K.~Khosoussi, M.~Giamou, J.~P. How, and J.~Kelly, ``Near-{Optimal}
  {Budgeted} {Data} {Exchange} for {Distributed} {Loop} {Closure}
  {Detection},'' \emph{arXiv:1806.00188 [cs]}, June 2018, arXiv: 1806.00188.

\bibitem{Tian2019}
Y.~Tian, K.~Khosoussi, and J.~P. How, ``A {Resource}-{Aware} {Approach} to
  {Collaborative} {Loop} {Closure} {Detection} with {Provable} {Performance}
  {Guarantees},'' \emph{arXiv:1907.04904 [cs]}, July 2019, arXiv: 1907.04904.

\bibitem{giamou_talk_2017}
M.~Giamou, K.~Khosoussi, and J.~P. How, ``Talk {Resource}-{Efficiently} to
  {Me}: {Optimal} {Communication} {Planning} for {Distributed} {Loop} {Closure}
  {Detection},'' \emph{arXiv:1709.06675 [cs]}, Sept. 2017, arXiv: 1709.06675.

\bibitem{Pinciroli2016}
C.~{Pinciroli} and G.~{Beltrame}, ``Buzz: An extensible programming language
  for heterogeneous swarm robotics,'' in \emph{2016 IEEE/RSJ International
  Conference on Intelligent Robots and Systems (IROS)}, Oct 2016, pp.
  3794--3800.

\bibitem{Labbe2019}
M.~Labbe and F.~Michaud, ``{RTAB-Map} as an open-source lidar and visual
  simultaneous localization and mapping library for large-scale and long-term
  online operation,'' \emph{Journal of Field Robotics}, vol.~36, no.~2, pp.
  416--446, 2019.

\bibitem{opencv_library}
G.~Bradski, ``{The OpenCV Library},'' \emph{Dr. Dobb's Journal of Software
  Tools}, 2000.

\bibitem{Smith87}
R.~Smith and P.~Cheeseman, ``On the representation and estimation of spatial
  uncertainty,'' \emph{Intl. J. of Robotics Research}, vol.~5, no.~4, pp.
  56--68, 1987.

\bibitem{DARPASubT}
DARPA, ``{DARPA Subterranean Challenge},''
  \url{https://www.subtchallenge.com/}, 2019, accessed: 2019-09-09.

\bibitem{Shi94}
J.~Shi and C.~Tomasi, ``Good features to track,'' in \emph{IEEE Conf. on
  Computer Vision and Pattern Recognition (CVPR)}, 1994, pp. 593--600.

\bibitem{rublee2011iccv-orb}
E.~Rublee, V.~Rabaud, K.~Konolige, and G.~Bradski, ``{ORB}: An efficient
  alternative to {SIFT} or {SURF},'' in \emph{Intl. Conf. on Computer Vision
  (ICCV)}.\hskip 1em plus 0.5em minus 0.4em\relax IEEE, 2011, pp. 2564--2571.

\bibitem{Dellaert12tr}
F.~Dellaert, ``Factor graphs and {GTSAM}: A hands-on introduction,'' Georgia
  Institute of Technology, Tech. Rep. GT-RIM-CP\&R-2012-002, September 2012.

\bibitem{Pinciroli2012}
C.~Pinciroli, V.~Trianni, R.~O'Grady, G.~Pini, A.~Brutschy, M.~Brambilla,
  N.~Mathews, E.~Ferrante, G.~{Di Caro}, F.~Ducatelle, M.~Birattari, L.~M.
  Gambardella, and M.~Dorigo, ``{ARGoS}: a modular, parallel, multi-engine
  simulator for multi-robot systems,'' \emph{Swarm Intelligence}, vol.~6,
  no.~4, pp. 271--295, 2012.

\bibitem{Lajoie19tr-DOORSLAM}
P.~Lajoie, B.~Ramtoula, Y.~Chang, L.~Carlone, and G.~Beltrame, ``{DOOR-SLAM}:
  Distributed, online, and outlier resilient {SLAM} for robotic teams,'' Tech.
  Rep., 2019, arXiv preprint: 1909.12198,
  \linkToPdf{https://arxiv.org/pdf/1909.12198.pdf}, Supplemental Material:
  \linkToPdf{https://www.dropbox.com/s/wgoqhiz8b96dl88/supplemental_material.pdf?dl=0}.

\end{thebibliography}

\end{document}